# IMPLEMENTING A GRU NEURAL NETWORK FOR FLOOD PREDICTION IN ASHLAND CITY, TENNESSEE.

George K. Fordjour[1] and Alfred J. Kalyanapu[1+]

[1] *Department of Civil and Environmental Engineering, Tennessee Technological University, Cookeville, TN 38505, USA*

[+] *Correspondence: akalyanapu@tntech.edu*

**ABSTRACT**

Ashland City, Tennessee, located within the Lower Cumberland Sycamore watershed, is highly susceptible to flooding due to increased upstream water levels. This study aimed to develop a robust flood prediction model for the city, utilizing water level data at 30-minute intervals from ten USGS gauge stations within the watershed. A Gated Recurrent Unit (GRU) network, known for its ability to effectively process sequential time-series data, was used. The model was trained, validated, and tested using a year-long dataset (January 2021-January 2022), and its performance was evaluated using statistical metrics including Nash-Sutcliffe Efficiency (NSE), Root Mean Squared Error (RMSE), Percent Bias (PBIAS), Mean Absolute Error (MAE), and Coefficient of Determination ($R^2$). The results demonstrated a high level of accuracy, with the model explaining 98.2% of the variance in the data. Despite minor discrepancies between predicted and observed values, the GRU model proved to be an effective tool for flood prediction in Ashland City, with potential applications for enhancing disaster preparedness and response efforts in Ashland City.

**1.0 INTRODUCTION**

The devastating impact of floods has intensified in both developed and developing countries, highlighting a growing global crisis (Ahern et al., 2005). The economic toll of floods extends beyond immediate damages, disrupting businesses and straining local resources for extended periods. Beyond causing extensive physical damage to infrastructure and property, floods also pose significant threats to public health through water contamination, soil liquefaction, and the potential for outbreaks of waterborne diseases (Atarigiya et al., 2017; Hirabayashi et al., 2013). In the United States, floods have led to loss of lives, and property damage. On average, each flooding event claims about 18 lives and incurs a weekly expenditure of around $75 million in dealing with flood-related issues (Smith, 2020). Outdated infrastructure and inadequate drainage systems in urban and semi-urban areas leave them particularly vulnerable to flooding (Brody et

al., 2011). Rivers in urban and semi-urban areas, like those in Nashville and Ashland City, are prone to rapid increases in water levels after heavy rainfall, often leading to flooding. The 2010 Nashville floods, driven by intense rains, caused the Cumberland River to rise swiftly, resulting in extensive damage, fatalities, and an estimated $1.5 billion in damages, highlighting the city's susceptibility to floods (Linscott et al., 2022). Similarly, the 2021 Ashland City floods, which claimed several lives, including a man in his 60s, further illustrate the serious impact of floods in urbanized areas (National Weather Service, 2021). These events emphasize the importance of accurately predicting river water levels to issue timely flood warnings, a crucial aspect of effective flood warning systems and water resource management (Li and Tan, 2015).

In the past decade, physically based models have played a significant role in predicting flood water levels (Borah, 2011). However, their reliance on extensive hydrological data can be a major obstacle, requiring specialized knowledge that poses a challenge for widespread use (Kim et al., 2015). Additionally, their short-term forecasting capabilities may be limited (Costabile and Macchione, 2015). In recent times, data-driven models have become increasingly favored for flood prediction. Their rise in popularity is partly due to their ability to numerically capture the non-linear nature of floods using only historical data, without necessitating an understanding of the physical processes involved. Moreover, these data-driven models, such as traditional machine learning (ML) techniques and deep learning (DL) models, are valued for their rapid development and effectiveness with minimal input requirements (Mosavi et al., 2018). However, while traditional ML techniques require specific feature engineering of raw data before processing, DL networks such as Recurrent Neural Networks (RNN) and Convolutional Neural Networks (CNN), have the advantage of autonomously identifying the necessary representations for detection or classification directly from the available data (Bentivoglio et al., 2022; LeCun et al., 2015). RNNs, particularly Long Short-Term Memory (LSTM) networks and Gated Recurrent Unit (GRU) networks, are highly effective in flood analysis, largely due to their capability in processing sequential data such as time-series (Aboah et al., 2023; Kratzert et al., 2019). GRU networks, in particular, offer a more computationally efficient alternative to LSTM networks, maintaining comparable performance with fewer gate computations (Kumar et al., 2023).

Therefore, the main aim of this study is to develop a robust and resource-efficient flood prediction model for Ashland City, Tennessee, utilizing water level data at 30-minute intervals. By leveraging the computational efficiency and proven capability of GRU networks in processing sequential time-series data, this model aims to enhance flood prediction accuracy, thereby contributing to timely disaster response and management efforts within the city.

**1.1 Related Work**

In recent years, GRU Neural networks have been successfully applied for various sequential tasks. Fu et al. (2016) utilized a GRU neural network for short-term traffic flow prediction and found it outperformed the LSTM in 84% of their time series analysis. Liu et al. (2017) a employed a GRU network to forecast China's primary energy consumption for 2021, highlighting its predictive capability. Zhang and Kabuka (2018) combined weather data with a GRU-based model for urban traffic flow prediction, achieving lower error rates than classical ML methods. Their GRU model achieved the lowest error when compared to classical ML techniques. Lastly, Le et al. (2019) developed a multi-layer GRU for identifying electron transport proteins, demonstrating its superior performance over existing predictors in this domain.

Furthermore, with respect to hydrological tasks, Sit and Demir (2019) applied GRU networks for decentralized flood forecasting in Iowa, where the model effectively predicted stage heights at various points. Similarly, Pan et al. (2020) combined GRU networks with a CNN model to forecast water levels in the Yangtze River, achieving better results compared to traditional ML algorithms such as autoregressive integrated moving average model (ARIMA) and wavelet-based artificial neural network (WANN). To further validate the effectiveness of GRU models, Zhou and Kang (2023) compared six popular ML models, including the GRU, for flood routing in the Yangtze River, with the GRU model performing better than the other ML models.

Building on these successful implementations of GRU networks in hydrological forecasting, our study aims to tailor this approach to the specific environmental and hydrological conditions of Ashland City, Tennessee. By integrating high-frequency water level data and leveraging the computational advantages of GRU networks, it is anticipated that the model will not only enhance the accuracy of flood predictions but also improve the efficiency of disaster response strategies.

**1.2 Study Area and Datasets**

    Ashland City, located in the Lower Cumberland Sycamore watershed (Hydrologic Unit Code 05130202) within the Cumberland River basin, serves as the county seat of Cheatham County. The city's significant susceptibility to flooding, primarily caused by increased water levels upstream, makes it an ideal location for this predictive study. The Lower Cumberland Sycamore watershed is well-equipped with numerous United States Geological Survey (USGS) gauge stations and is rich in historical data. Thus, water level data was collected from these gauge stations for the study. Figure 1 illustrates the study area, highlighting the ten USGS gauge stations within the Lower Cumberland Sycamore watershed that were used for data collection. The Ashland City gauge station, highlighted in green in Figure 1, served as the prediction point for this flood prediction analysis. The datasets from the stations were collected at 30-minute intervals. The data collected spanned the full year, from January 1, 2021, 00:00 to January 1, 2022, 00:00.

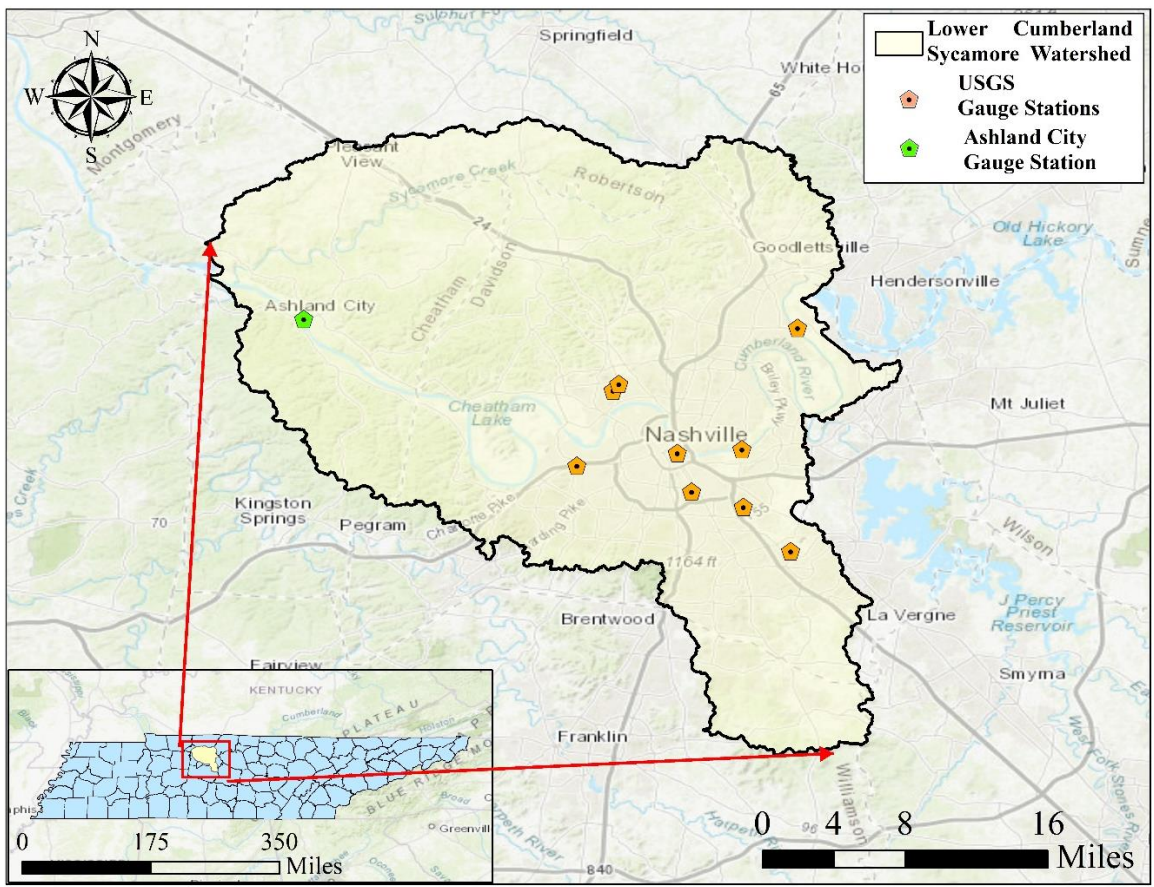

Figure 1: Lower Cumberland Sycamore Watershed with USGS gauge stations

The collected dataset was then split into three sets: 70% for training, 20% for validation, and 10% for testing. Table 1 outlines the specific time periods allocated for training, validation, and testing of the GRU model.

Table 1: Description of datasets used for model development

| Dataset | Start Date and Time (mm-dd-yyyy) | End Date and Time (mm-dd-yyyy) |
|---|---|---|
| Training Data | 01/01/2021, 00:00 | 08/18/2021, 22:00 |
| Validation Data | 08/18/2021, 22:30 | 11/25/2021, 11:30 |
| Testing Data | 11/25/2021, 12:00 | 01/01/2022, 00:00 |

The datasets were preprocessed for the model analysis. The linear interpolation method was used to fill data gaps. The predictor variables were scaled and transformed to ensure optimal performance of the GRU model. This approach aimed to improve the accuracy of flood predictions for Ashland City.

## 2.0 METHODOLOGY

This section outlines the methodology used for this study. It includes an explanation of the GRU neural network and the model design used for the analysis. Additionally, the evaluation metrics selected for this study are discussed in this section.

### 2.1 Gated Recurrent Unit (GRU) Neural Network

The Gated Recurrent Unit (GRU) is a type of Recurrent Neural Network (RNN) introduced by Cho et al. (2014). It addresses the vanishing gradient problem, which is a common issue in standard RNNs when handling long data sequences (Aboah and Arthur, 2021; Cho et al., 2014). Although GRUs are similar to LSTM networks, they have a simpler architecture with fewer parameters (Chung et al., 2014; Gao and Glowacka, 2016). GRU networks utilize two primary gates to manage information flow: the update gate and the reset gate. The update gate selectively determines how much of the previous hidden state's information should be retained for the current step. It filters out irrelevant details and helps the model focus on crucial patterns and long-term dependencies. Meanwhile, the reset gate controls the degree to which the previous hidden state is reset before incorporating new input information. This allows the model to adapt to changes in the input sequence, preventing the accumulation of irrelevant information (Dey and Salemt, 2017). By dynamically adjusting these gates, GRU networks effectively capture both short-term and long-term dependencies in sequential data, making them well-suited for time-series forecasting tasks like flood prediction. The equations that express the operation of a GRU cell are shown in equations 1-4:

$$U_t = \emptyset(V_1 \cdot X_t + V_2 \cdot H_{prev} + c_u) \tag{1}$$

$$R_t = \emptyset(V_3 \cdot X_t + V_4 \cdot H_{prev} + c_r) \tag{2}$$

$$H_t^* = \tanh(V_5 \cdot X_t + V_6(R_t \circ H_{prev}) + c_h) \tag{3}$$

$$H_t = U_t \circ H_{prev} + (1 - U_t) \circ H_t^* \tag{4}$$

Where $U_t$, $R_t$, $H_t^*$, and $H_t$ represent the update gate vector, reset gate vector, and candidate hidden state, and updated hidden state respectively all at time $t$. The hyperbolic tangent is represented by $\tanh$ and the sigmoid function is represented by $\emptyset$. $V$, $c$, and $X_t$ are the weight matrices, bias terms, and input water level at time $t$ respectively.

### 2.2 Model Design

The GRU model begins with an input layer that receives the water level time series data, formatted into sequences to capture the inherent temporal relationships. This sequential data is then passed to the GRU layer, which processes it across time steps and features, extracting meaningful patterns and dependencies. The GRU layer is configured to return sequences, ensuring that the temporal information is preserved and utilized effectively. Subsequently, a hidden dense layer with ReLU (Rectified Linear Unit) activation is used. This layer introduces non-linearity into the model, allowing it to learn complex relationships between the extracted features and the target variable. Following this, another hidden dense layer, also with ReLU activation, is added to further increase the model's capacity to learn intricate patterns in the data. Finally, an output dense layer is used to generate the predicted water level values. Table 2 outlines the training parameters used for building the model. These parameters were selected using the random search strategy (Bergstra and Bengio, 2012). The model uses the Adam optimizer and mean squared error as its loss function.

Table 2: Training Parameters used to build the GRU model

| Training Parameter | Parameter Value |
|---|---|
| GRU layer | 128 |
| Dense layer 1 | 160 |
| Dense layer 2 | 96 |
| Output dense layer | 1 |
| Learning rate | 0.0001 |
| Epochs | 30 |

To predict the next water level, the GRU model was designed to utilize an input sequence of 5 hours, equivalent to 10 timesteps of 30-minute interval water level data. This allowed the model to consider recent historical trends when making predictions. The model was trained on the training dataset, validated using the validation dataset, and its final performance was evaluated on the unseen test dataset. The model design used in this study is illustrated in Figure 2.

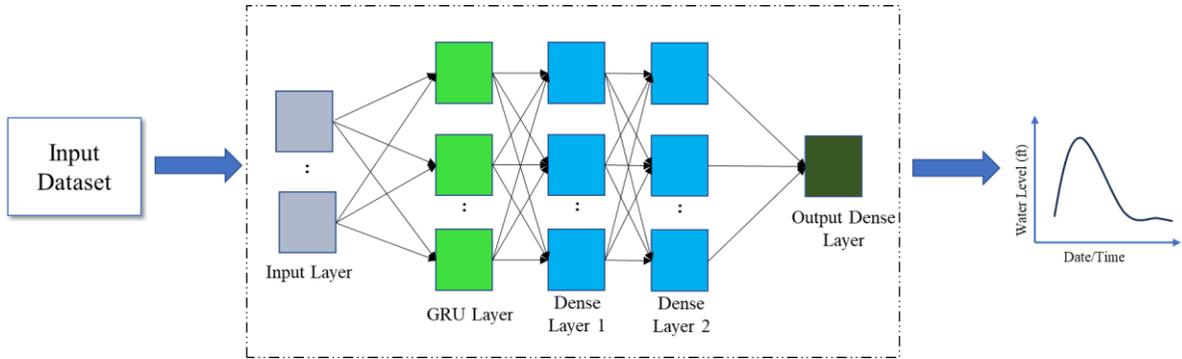

Figure 2: GRU model design used for analysis

## 2.3 Model Performance Evaluation

Hydrological prediction models are usually evaluated using a variety of statistical metrics (Moriasi et al., 2015). In this study, the metrics used to evaluate the performance of the GRU model includes: Nash-Sutcliffe Efficiency (NSE), Root Mean Squared Error (RMSE), Percent Bias (PBIAS), Mean Absolute Error (MAE), and Coefficient of Determination ($R^2$). The aforementioned metrics were computed using the hydrostats Python package (Owusu et al., 2022; Roberts et al., 2018), a specialized tool for hydrological analysis and model evaluation. The equations for these statistical metrics (NSE, RMSE, PBIAS, MAE, and $R^2$) are detailed in the equations 5-9.

$$NSE = 1 - \frac{\sum_{k=1}^{N}(o_k - p_k)^2}{\sum_{k=1}^{N}(o_k - \bar{o})^2} \tag{5}$$

$$RMSE = \sqrt{\frac{1}{N} \sum_{k=1}^{N}(o_k - p_k)^2} \qquad (6)$$

$$PBIAS = \frac{\sum_{k=1}^{N}(o_k - p_k) \times 100}{\sum_{k=1}^{N} o_k} \qquad (7)$$

$$MAE = \frac{1}{N}\sum_{k=1}^{N}|o_k - p_k| \qquad (8)$$

$$R^2 = \left[\frac{\sum_{k=1}^{N}(o_k-\bar{o})(p_k-\bar{p})}{\sqrt{\sum_{k}^{N}(o_k-\bar{o})^2}\sqrt{\sum_{k}^{N}(p_k-\bar{p})^2}}\right]^2 \qquad (9)$$

## 3.0 RESULTS AND DISCUSSION

The performance of the GRU model was assessed using metrics outlined in section 2.3. The performance results based on the different statistical metrics are detailed in Table 3.

Table 3: Statistical performance of model predictions

| Evaluation Metric | Performance Value |
|---|---|
| RMSE (ft) | 0.064 |
| NSE | 0.982 |
| PBIAS (%) | 0.068 |
| $R^2$ | 0.984 |
| MAE (ft) | 0.052 |

As detailed in Table 3, the GRU model achieved an RMSE of 0.064ft and a MAE of 0.052ft, indicating a strong model performance. These values, approaching zero, suggest a close match between predicted and observed water levels, aligning with established benchmarks for acceptable model accuracy (Moriasi et al., 2007; Singh et al., 2005). Specifically, the RMSE value suggests a minimal average magnitude of prediction errors, while the MAE metric demonstrates a very small average absolute error between the model's predictions and the actual data points. The NSE value of 0.982 indicates that the model explains 98.2% of the variance around the mean, highlighting its ability to capture the observed patterns effectively. This is further supported by the $R^2$ value of 0.984, signifying a very good correlation between the predicted and observed values across the entire dataset. Additionally, the PBIAS value of 0.068% indicates a slight underestimation of the observed values, which falls within the "very good" accuracy range as defined by (Moriasi et al., 2007), as a value within ±10% is considered very good. These results collectively affirm the GRU model's strong performance in predicting flood events in Ashland City.

Additionally, a visual comparison of the observed and predicted water levels is illustrated in Figure 3. The observed water levels are represented in blue while the predicted water levels shown in red. The model appears to closely track the observed water levels throughout the period. The predicted values closely follow the trends and fluctuations of the observed data, suggesting a good fit. The model successfully captures most of the peaks and troughs in the data. This is evident in several instances where both lines rise and fall in concert, such as around mid-December and late December.

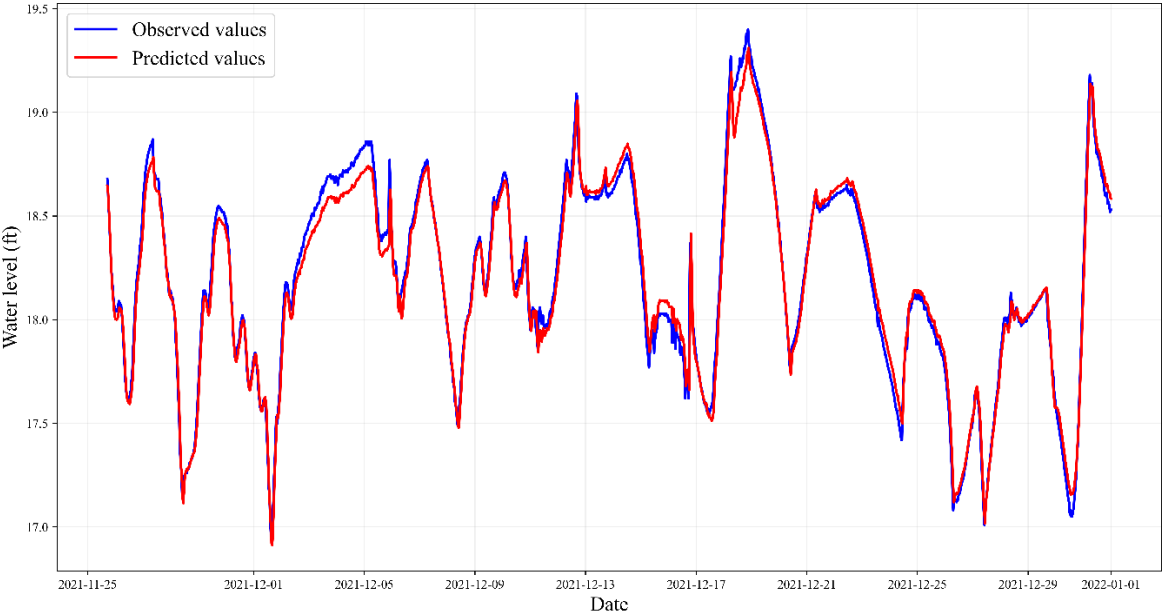

Figure 3: Comparison of observed and predicted water levels at Ashland City

While the overall model fit is quite good, there are a few instances where the predicted where predicted values slightly deviate from the observed data, particularly in early December (Figure 3). These instances could potentially be improved with further model refinement, such as adjusting learning rates, or increasing training iterations.

## 4.0 CONCLUSION

In conclusion, this study demonstrates the successful application of a GRU network for flood prediction in Ashland City, Tennessee. By leveraging 30-minute interval water level data from USGS gauge stations within the Lower Cumberland Sycamore watershed, the model achieved a high level of accuracy, as evidenced by the statistical performance metrics. Although the model demonstrated a good fit overall, some slight discrepancies between predicted and observed values were observed, particularly in early December. Despite these minor inconsistencies, the GRU network's ability to effectively capture complex temporal patterns and dependencies in hydrological data was shown. This research highlights the potential of GRU-based models as valuable tools for flood forecasting and early warning systems, ultimately contributing to more effective disaster preparedness and response strategies in flood-prone areas like Ashland City. Future research could explore further model refinements, such as incorporating additional environmental variables such as rainfall, to further enhance prediction accuracy and address the minor discrepancies observed in specific time periods.

## ACKNOWLEDGEMENTS

The authors extend their gratitude to the Center of Management, Utilization, and Protection of Water Resources at Tennessee Technological University for their financial support. The authors are also grateful to Dr. Mattingly Hayden and Dr. Mark Green for their valuable input in reviewing this work.

## DATA AVAILABILITY

The data for this study are available and can be shared upon reasonable request from the corresponding author.